\documentclass[runningheads]{llncs}

\usepackage{eccv}

\usepackage{graphicx}

\usepackage{hyperref}

\titlerunning{CodeSCAN}

\author{Alexander Naumann \and
    Felix Hertlein \and
    Jacqueline Höllig \and
    Lucas Cazzonelli \and 
    Steffen Thoma
    }

\authorrunning{A. Naumann et al.}

\institute{FZI Research Center for Information Technology, Karlsruhe, Germany \\
Karlsruhe Institute of Technology (KIT), Karlsruhe, Germany
    \email{\{anaumann,hertlein,hoellig,cazzonelli,thoma\}@fzi.de}}

\title{CodeSCAN: ScreenCast ANalysis for Video Programming Tutorials
}

\usepackage{booktabs}
\usepackage{wrapfig}

\usepackage[all]{nowidow} %
\usepackage{amsfonts}
\usepackage{amssymb}
\usepackage{graphicx}
\usepackage{caption} %
\usepackage{subcaption} %
\usepackage{enumitem}
\usepackage{color}
\usepackage{calc}
\usepackage{float}
\usepackage{bookmark}

\usepackage[style=ieee,
    doi=false,
    url=false,
    isbn=false,
    mincitenames=1,
    maxcitenames=1,
    minbibnames=6,
    maxbibnames=6,
    backend=biber]{biblatex}  %

\DeclareSourcemap{
    \maps[datatype=bibtex, overwrite]{
        \map{
            \step[fieldset=address, null]
            \step[fieldset=location, null]
            \step[fieldset=note, null]  %
            \step[fieldset=series, null]
            \step[fieldset=issn, null]
            \step[fieldset=pages, null]
            \step[fieldset=editor, null]
        }
    }
}

\newcommand{\eg}{e.g.\ } %

\newcommand{\codescan}{CodeSCAN}
\newcommand{\vscode}{Visual Studio Code}
\newcommand{\ghdev}{\href{https://github.dev}{https://github.dev}}
\newcommand{\gh}{Github}

\usepackage[capitalize]{cleveref}
\crefname{section}{Sec.}{Secs.}
\Crefname{section}{Section}{Sections}
\Crefname{table}{Table}{Tables}
\crefname{table}{Tab.}{Tabs.}
\begin{document}
\title{
    CodeSCAN: ScreenCast ANalysis for Video Programming Tutorials
}
\maketitle              %

\begin{abstract}
    Programming tutorials in the form of coding screencasts play a crucial role in programming education, serving both novices and experienced developers.
    However, the video format of these tutorials presents a challenge due to the difficulty of searching for and within videos.
    Addressing the absence of large-scale and diverse datasets for screencast analysis, we introduce the \codescan{} dataset.
    It comprises 12,000 screenshots captured from the \vscode{} environment during development, featuring 24 programming languages, 25 fonts, and over 90 distinct themes, in addition to diverse layout changes and realistic user interactions.
    Moreover, we conduct detailed quantitative and qualitative evaluations to benchmark the performance of Integrated Development Environment (IDE) element detection, color-to-black-and-white conversion, and Optical Character Recognition (OCR).
    We hope that our contributions facilitate more research in coding screencast analysis, and we make the source code for creating the dataset and the benchmark publicly available at \href{https://a-nau.github.io/codescan}{a-nau.github.io/codescan}.
\end{abstract}

\section{Introduction}

In the ever-evolving landscape of programming education and knowledge dissemination, coding screencasts on platforms such as YouTube have emerged as powerful tools for both novice and experienced developers.
These video tutorials not only provide a visual walkthrough of coding processes but also offer a unique opportunity for learners to witness real-time problem-solving, coding techniques, and best practices.
As the popularity of coding screencasts continues to soar, the possibility to augment traditional video content with additional information to improve the learner's experience, becomes increasingly interesting and relevant.
The source code extracted from a screencast, that replicates the full coding project up to the given position in the video, is one such element that can empower learners with the ability to delve deeper into the presented material.

\begin{figure}
    \centering
    \includegraphics[width=0.65\linewidth]{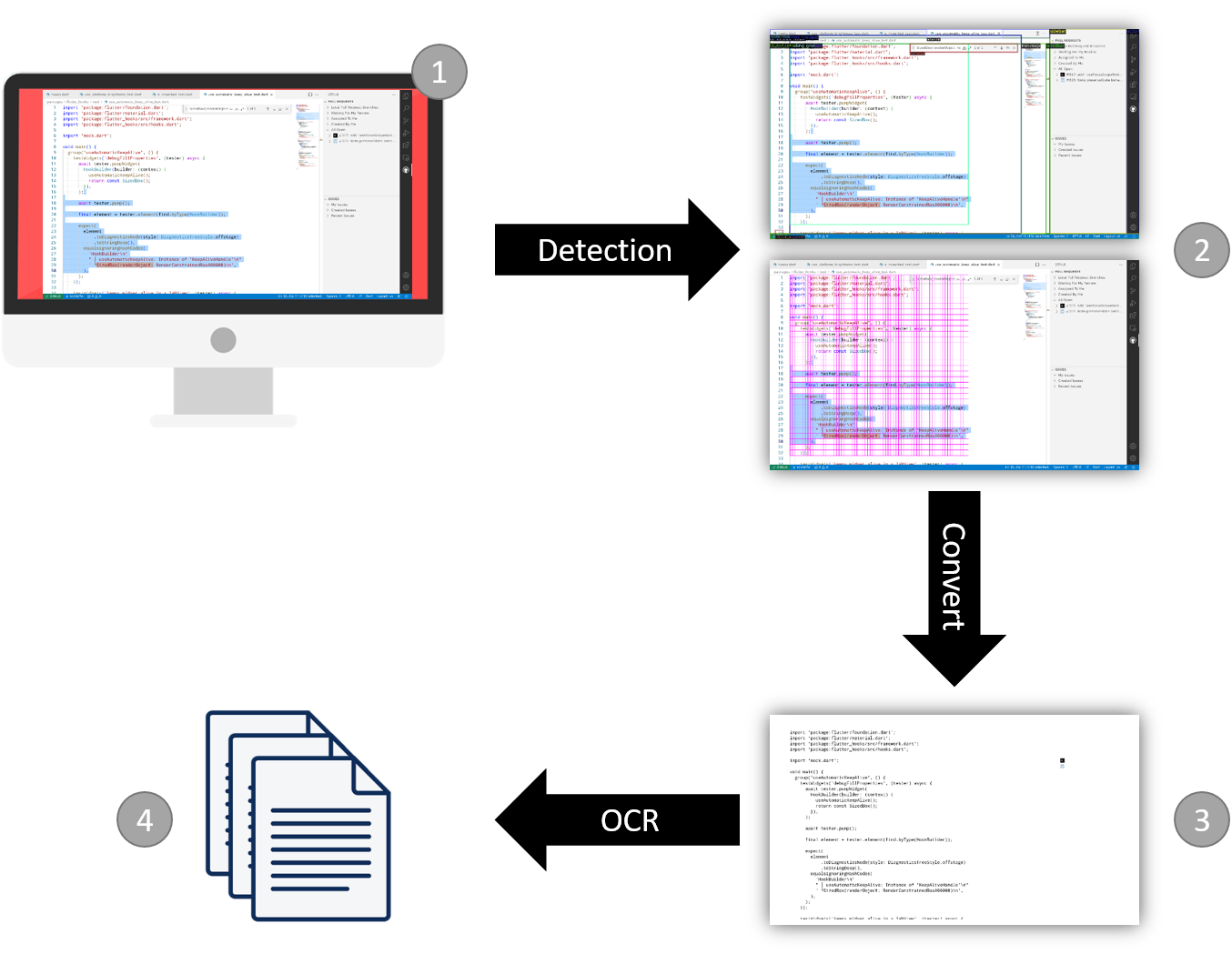}
    \caption{
        Source code extraction pipeline:
        Given an image from a coding screencast (1), the IDE elements need to be detected (2).
        This is followed by an optional binarization step (3) which converts the color image into a black-and-white image, and finally OCR is applied (4) to read out the source code.
    }
    \label{fig:overview}
    \vspace{-0.5cm}
\end{figure}

Incorporating such code extraction into the realm of coding screencasts enables learners to benefit from a dual advantage.
Firstly, it empowers video search algorithms to utilize the full source code, enhancing the precision and relevance of search results.
By indexing and analyzing the extracted code, search engines can pinpoint specific programming concepts, language features, or problem-solving techniques, leading learners to the most pertinent videos that align with their educational needs.
Rather than watching an entire video in search of a specific code segment, learners can navigate directly to the relevant sections, streamlining the learning process and increasing overall comprehension.
Secondly, having access to the full code at any given point in a coding screencast also empowers learners to engage in hands-on experimentation seamlessly.
By providing learners with the entire codebase in its current state, irrespective of their position in the video timeline, the educational experience is transformed into an interactive playground for exploration and experimentation, where the code can be executed and experimented with at any stage within the video.

However, the task of retrieving the full source code of a project at any given point in a coding screencast is challenging: Instructors frequently navigate between different files within a project, introduce small modifications or copy-paste large amounts of code and additionally, the character distribution of source code differs strongly from standard text paragraphs.
To cope with this it is crucial to understand the visible IDE components in addition to being able to extract the visible source code as text.
To this end, we present a novel large-scale and high-quality dataset for coding screencast analysis called \codescan{}, in this work.
It contains more than 12,000 screenshots of coding projects in 24 different programming languages.
These screenshots are independent and cannot be composed into coherent videos.
While full programming tutorials would enable better end-to-end testing of the pipeline, it is important to note, that all current approaches only use single frames as input.
\codescan{} is the first dataset of its kind that sets itself apart by its unprecedented size, diversity and annotation granularity.
We use \vscode{} with approximately 100 different themes and 25 different fonts.
Through automated interaction with \vscode{}, we achieve large visual diversity by \eg{} editing code, highlighting areas, performing search, and much more.
In addition to the dataset, we present a detailed literature review and analyze the performance for text recognition on integrated development environment (IDE) images in detail.
More specifically, we analyze the influence of different OCR engines, image binarization and image quality.
An overview of a potential source code extraction pipeline is outlined in \cref{fig:overview}.

To summarize, the main contributions of our work are

\begin{itemize}
    \item we introduce \codescan{}, a novel high-quality dataset for screencast analysis of video programming tutorials with 12,000 screenshots containing a multiplicity of annotations and high diversity regarding programming languages and visual appearance,
    \item we evaluate a baseline CNN on diverse IDE element detection and propose the usage of a so-called \textit{coding grid} for source code localization,
    \item we benchmark five different OCR engines, and analyze the influence of image binarization and image quality on performance.
\end{itemize}

The remainder of this work is organized as follows.
We present an overview of related work in \cref{sec:rel_work}.
Subsequently, we present the details of our dataset generation approach in \cref{sec:data}.
\cref{sec:eval} presents the evaluation for IDE element detection and OCR.
Finally, \cref{sec:conclusion} concludes the paper.

\section{Related Work}
\label{sec:rel_work}

We review the existing literature on approaches for code extraction, tools using code extraction, and, finally, available datasets.

\subsection{Approaches}
A common pipeline to extract the source code from a programming screencast as text looks as follows: (1) the video is split up into images and duplicates are removed, (2) the image is classified as containing or not containing source code, (3) the source code is located by predicting a bounding box, and finally (4) OCR is applied within this bounding box to extract the source code as text.
This full pipeline is necessary to be able to perform inference on new, unseen video programming tutorials.
However, the literature frequently focuses only on a subset of these tasks.
\textcite{ottDeepLearningApproach2018} trained a CNN classifier to identify whether a screenshot contains fully visible code, partially visible code, handwritten code or no code.
\textcite{ottLearningLexicalFeatures2018} use a CNN to locate the code and to classify its programming language (Java or Python) purely by using image features.
\textcite{alahmadiCodeLocalizationProgramming2020} tackle code localization to remove the noise in OCR results which is introduced by additional IDE elements such as the menu by comparing five different backbone architectures.
\textcite{malkadiStudyAccuracyOCR2020} focus on comparing different OCR engines, thus, focusing on step (3).
The literature analyzes between one and seven different programming languages (if these are specified).
\textit{Java} is most popular \cite{yadidExtractingCodeProgramming2016,ponzanelliCodeTubeExtractingRelevant2016,ponzanelliTooLongDidn2016,ottDeepLearningApproach2018,alahmadiAccuratelyPredictingLocation2018,ottLearningLexicalFeatures2018,zhaoActionNetVisionBasedWorkflow2019,ponzanelliAutomaticIdentificationClassification2019,baoVTRevolutionInteractiveProgramming2019,berghCuratedSetLabeled2020,baoEnhancingDeveloperInteractions2020,baoPsc2codeDenoisingCode2020,alahmadiCodeLocalizationProgramming2020},
followed by \textit{Python} \cite{ottLearningLexicalFeatures2018,zhaoActionNetVisionBasedWorkflow2019,berghCuratedSetLabeled2020,alahmadiCodeLocalizationProgramming2020,malkadiStudyAccuracyOCR2020},
\textit{C\#} \cite{alahmadiCodeLocalizationProgramming2020,malkadiStudyAccuracyOCR2020}
and
\textit{XML} \cite{alahmadiVID2XMLAutomaticExtraction2023}.
Also, different IDEs, such as
Eclipse \cite{baoReverseEngineeringTimeseries2015,baoExtractingAnalyzingTimeseries2017,baoVTRevolutionInteractiveProgramming2019},
Visual Studio \cite{malkadiStudyAccuracyOCR2020}
or a diverse mix of multiple IDEs \cite{ottDeepLearningApproach2018,alahmadiAccuratelyPredictingLocation2018,ottLearningLexicalFeatures2018,zhaoActionNetVisionBasedWorkflow2019,alahmadiCodeLocalizationProgramming2020,alahmadiUIScreensIdentification2020,malkadiStudyAccuracyOCR2020,alahmadiVID2XMLAutomaticExtraction2023}
are used.
Other related tasks include recognizing user actions (e.g. enter or select text) in coding screencasts \cite{zhaoActionNetVisionBasedWorkflow2019} and identifying UI screens in videos \cite{alahmadiUIScreensIdentification2020}.

\subsection{Tools}

\textcite{ponzanelliCodeTubeExtractingRelevant2016} present CodeTube, a recommender system that leverages textual, visual and audio information from coding screencasts.
The users inputs a query, and CodeTube intends to return a relevant, cohesive and self-contained video.
\textcite{baoPsc2codeDenoisingCode2020} introduce ps2code, a tool which leverages code extraction to provide a coding screencast search engine and a screencast watching tool which enables interaction.
PSFinder \cite{yangEfficientSearchLiveCoding2022} classifies screenshots into containing or not containing an IDE to identify live-coding screencasts.

\subsection{Datasets}
To the best of our knowledge, only two of the previously mentioned works have their full annotated datasets publicly available.
The dataset presented by \textcite{malkadiStudyAccuracyOCR2020} comprises screenshots of only the visible code with the associated source code files in Java, Python and C\#.
The dataset was created manually, and no pixel-wise correspondence between source code and images is available.
Furthermore, there is low diversity in the IDE color scheme and font.
\textcite{alahmadiUIScreensIdentification2020} present a dataset where UI screenshots are annotated with bounding boxes.
Since they do not target code extraction, no such annotations are provided.

\subsection{Discussion}
To empower learners to engage in hands-on experimentation seamlessly and to improve the search quality within and across videos, it is crucial to be able to extract the source code of a complete project.
This requires, for example, to identify the file tree and the currently active tab.
Moreover, it is important to reliably and incrementally edit existing files.
In order to enable such sophisticated full-project source code extraction, it is essential to have a large, high-quality dataset with detailed annotations.
Since currently only one relevant annotated dataset with insufficient annotation granularity is publicly available \cite{malkadiStudyAccuracyOCR2020}, we present the novel dataset \codescan{}.
\codescan{} covers 24 different programming languages, over 90 \vscode{} themes and 25 different fonts.
We provide a multiplicity of automatically annotated pixel-accurate annotations that reach far beyond code localization.
Furthermore, we are the first to benchmark diverse IDE element localization and to analyze the influence of image binarization and image quality on OCR.

\section{Dataset Generation}
\label{sec:data}

Our dataset was acquired by scraping data from \ghdev{}.
The details on the dataset acquisition will be presented in \cref{sec:data:scrape}.
Subsequently, we lay out our annotation generation approach in \cref{sec:data:annos}.

\subsection{Data Acquisition}
\label{sec:data:scrape}

For the data acquisition, we exploit the fact that any \gh{} repository can be opened with a browser version of \vscode{} by changing the URL from
\begin{itemize}
    \item \textit{https://github.com/USER-NAME/REPOSITORY-NAME}
    \item to \textit{https://github.\textbf{dev}/USER-NAME/REPOSITORY-NAME}.
\end{itemize}
Thus, we randomly select $100$ repositories with a permissive license (MIT, BSD-3 or WTFPL) per programming language.
Since we consider $24$ programming languages, this results in a total of $2,400$ repositories.
The considered programming languages are
\textit{C},
\textit{C\#},
\textit{C++},
\textit{CoffeeScript},
\textit{CSS},
\textit{Dart},
\textit{Elixir},
\textit{Go},
\textit{Groovy},
\textit{HTML},
\textit{Java},
\textit{JavaScript},
\textit{Kotlin},
\textit{Objective-C},
\textit{Perl},
\textit{PHP},
\textit{PowerShell},
\textit{Python},
\textit{Ruby},
\textit{Rust},
\textit{Scala},
\textit{Shell},
\textit{Swift,} and
\textit{TypeScript}.
We select five files with file endings corresponding uniquely to the underlying language (e.g. \textit{.py} for \textit{Python}) at random per repository, leading to a total of $12,000$ files.
To achieve highly diverse IDE appearances, we use more than $90$ different \vscode{} themes and $25$ monospaced fonts.
Moreover we apply the following fully automated, random layout changes
\begin{itemize}
    \item different window sizes (for desktop $1920\times1200$, $1920\times1080$, $1200\times1920$, $1440\times900$ and tablet $768\times1024$, $1280\times800$, $800\times1280$),
    \item layout type (classic or centered layout),
    \item coding window width,
    \item degree of zoom for the source code text,
    \item output panel visibility, size and type (e.g. Terminal, Debugger, etc.),
    \item sidebar visibility, location, size and type (toggling sidebar options),
    \item menubar visibility (classic, compact, hidden, visible),
    \item activity bar visibility,
    \item status bar visibility,
    \item breadcrumbs visibility,
    \item minimap visibility, and
    \item control characters and white spaces rendering visibility
\end{itemize}
and realistic user interactions
\begin{itemize}
    \item highlighting multiple lines of code,
    \item right clicking at random positions in the code,
    \item adding or removing characters from the source code,
    \item search within the opened file,
    \item search within the project, and
    \item open a random number of other tabs.
\end{itemize}

The automation of all these appearance changes was implemented using the framework Selenium\footnote{See \href{https://www.selenium.dev/documentation/}{https://www.selenium.dev/documentation/}.}.
We refer to one such final IDE configuration (i.e. the full visual appearance of the IDE with the current active source code file and its appearance characteristics) as \textit{scene} in the following.
To persist the current scene, we export the scene webpage using SingleFile\footnote{See \href{ https://github.com/gildas-lormeau/SingleFile}{ https://github.com/gildas-lormeau/SingleFile}.}.
In addition, we save the full source code of the currently visible file as reference.
Thus, at the end of the dataset acquisition phase, we have a \textit{source.txt} and \textit{page.html} file for each of the $12,000$ code files.

Note, that especially selecting random IDE themes is challenging to automate\footnote{This is due to strongly varying loading times of the theme and since themes frequently have multiple available color schemes.} and thus, not always the desired theme was applied correctly for each scene.
Since each scene is represented in a single HTML file, however, we are able to leverage the CSS specifications (background and font color) to perform theme-wise clustering after the data acquisition for quality assurance.

\subsection{Annotation Generation}
\label{sec:data:annos}

\begin{figure}
    \begin{subfigure}[b]{0.48\linewidth}
        \centering
        \fbox{\includegraphics[width=\linewidth]{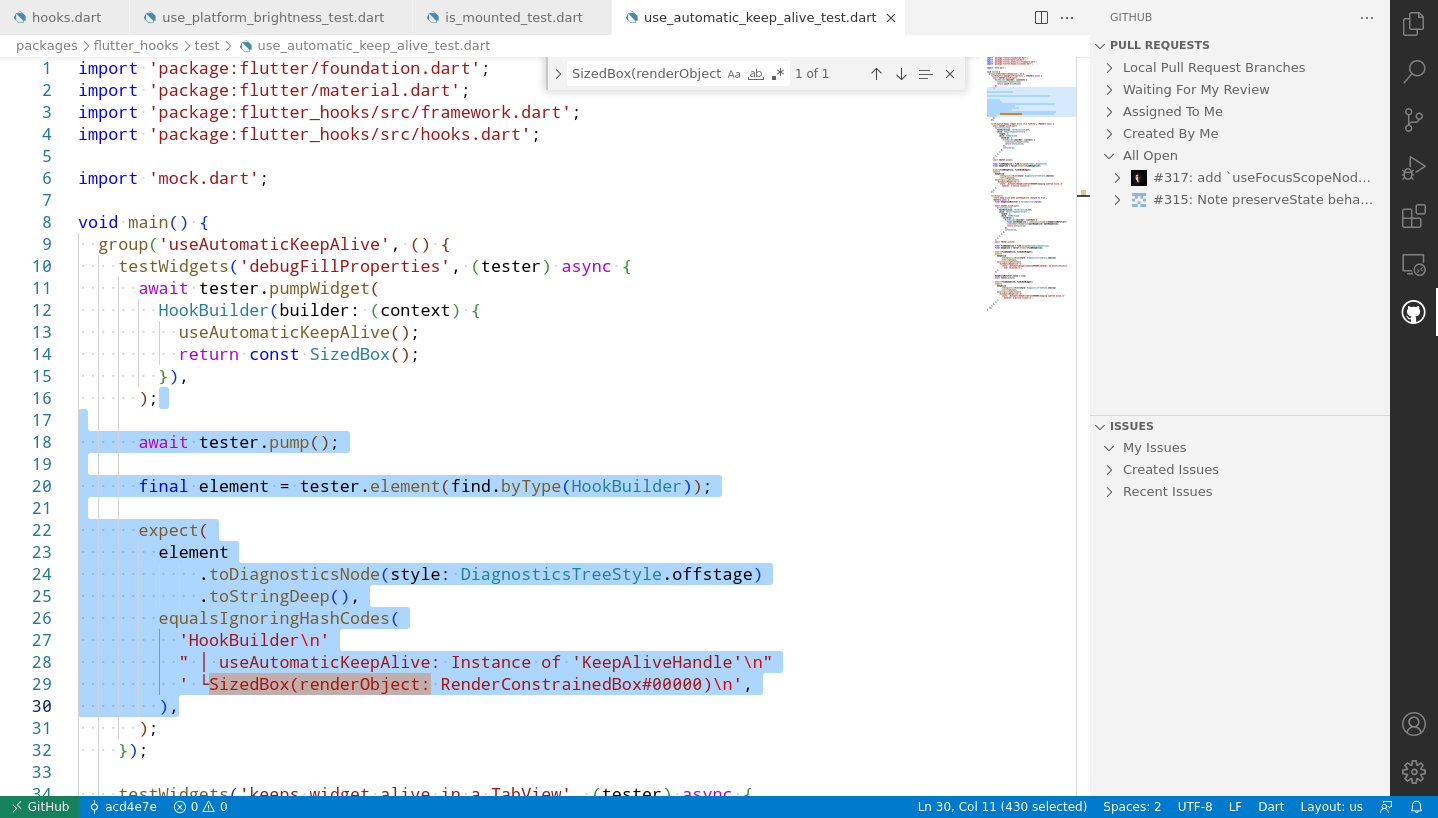}}
        \caption{Color Image of Scene}
        \label{fig:data:types:color}
        \vspace{2ex}
    \end{subfigure}%
    \hspace{3mm}%
    \begin{subfigure}[b]{0.48\linewidth}
        \centering
        \fbox{\includegraphics[width=\linewidth]{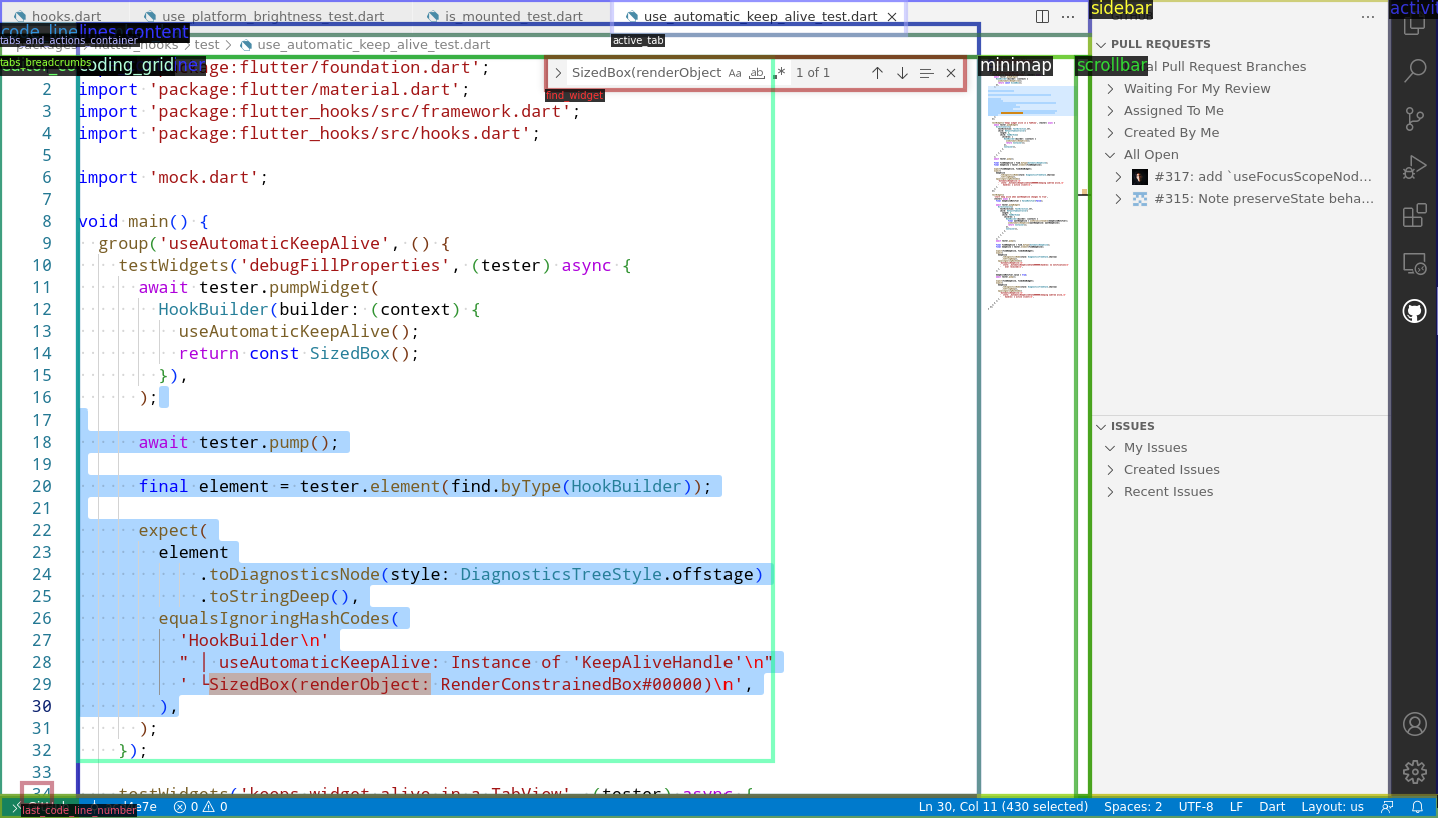}}
        \caption{Bounding Box Annotations}
        \label{fig:data:types:bbox}
        \vspace{2ex}
    \end{subfigure}
    \begin{subfigure}[b]{0.48\linewidth}
        \centering
        \fbox{\includegraphics[width=\linewidth]{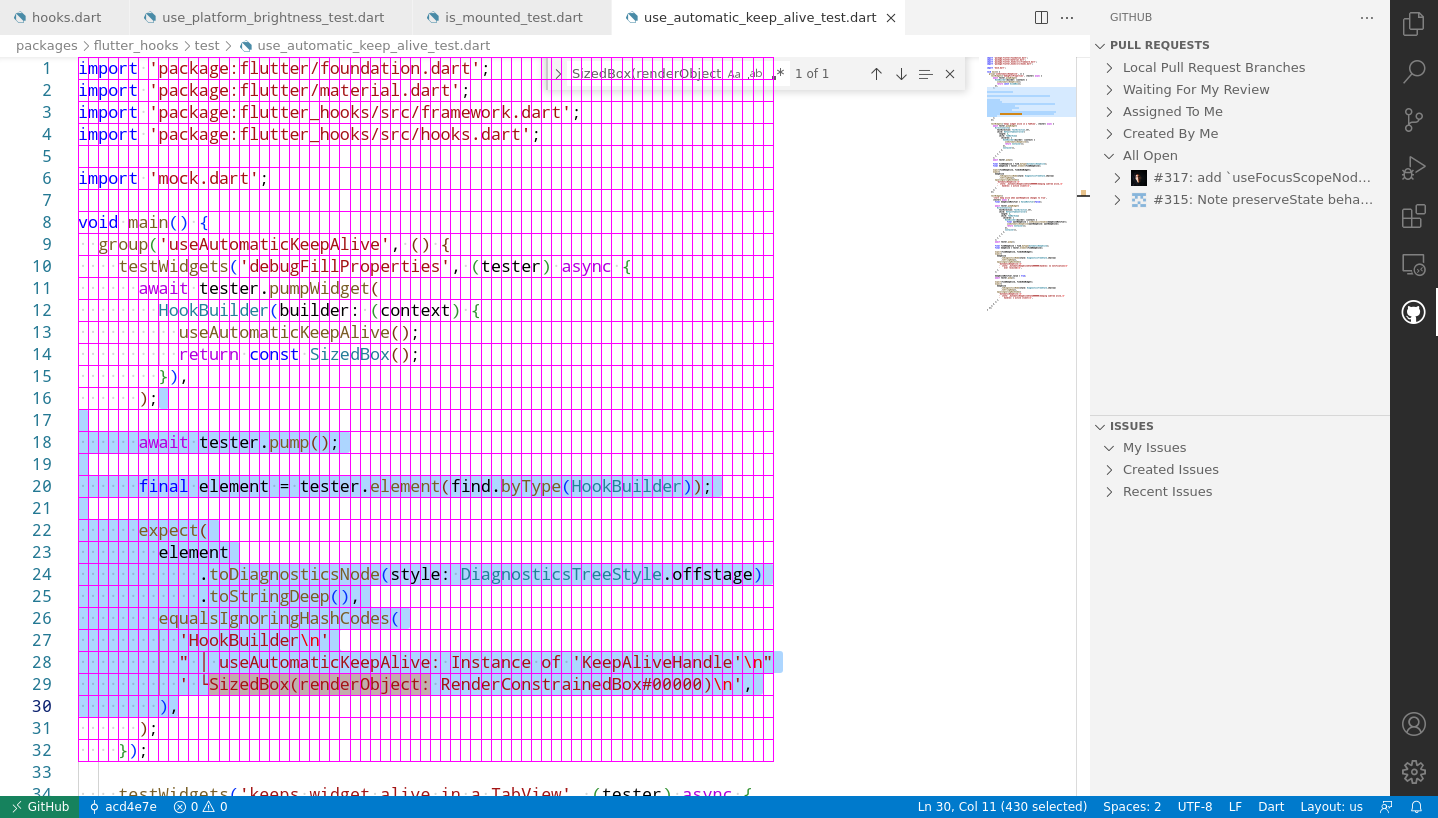}}
        \caption{Coding Grid Annotation}
        \label{fig:data:types:grid}
        \vspace{2ex}
    \end{subfigure}%
    \hspace{3mm}%
    \begin{subfigure}[b]{0.48\linewidth}
        \centering
        \fbox{\includegraphics[width=\linewidth]{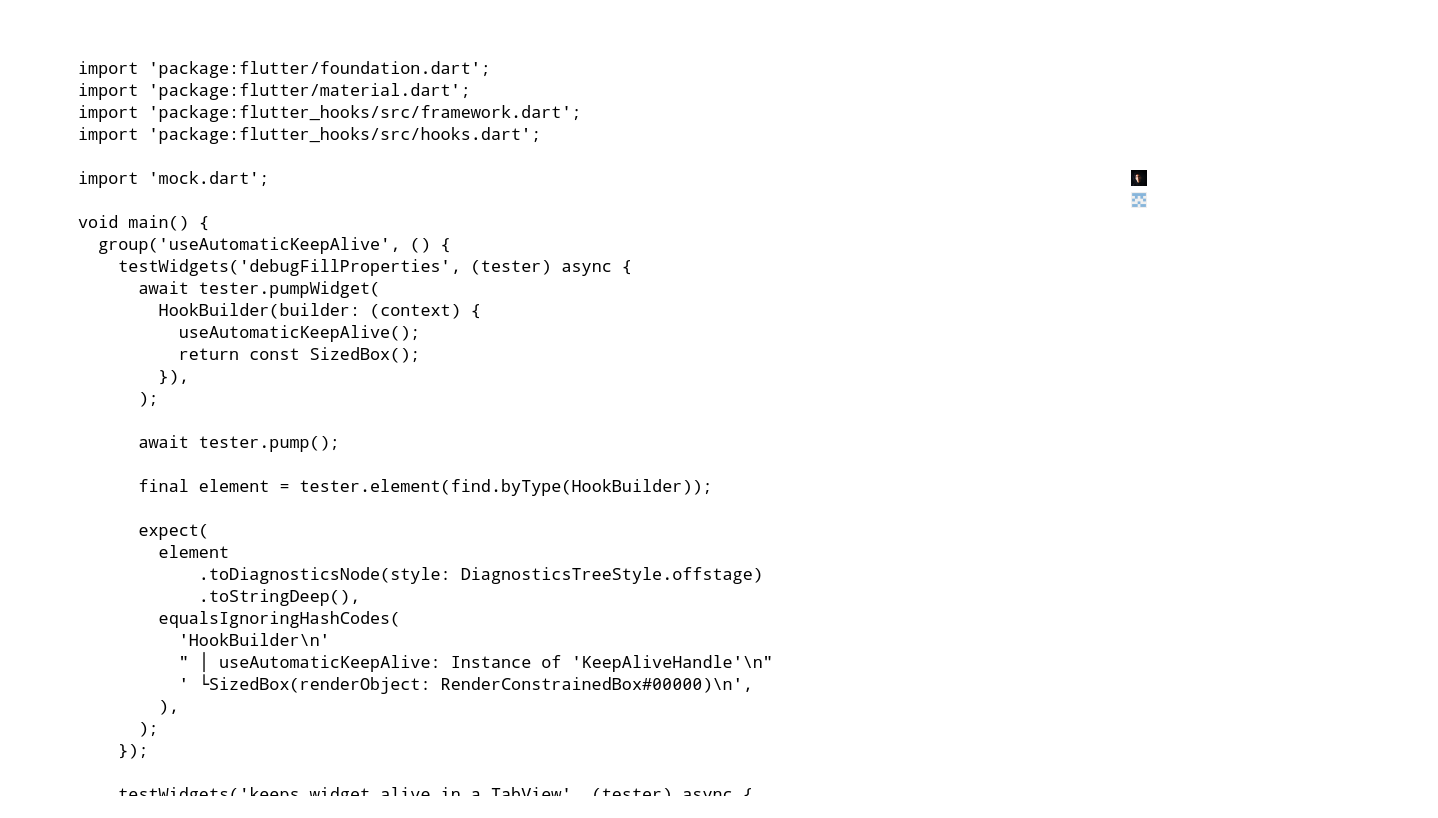}}
        \caption{Black-and-white Image of Scene}
        \label{fig:data:types:bw}
        \vspace{2ex}
    \end{subfigure}
    \caption{
        Example of the different available annotation and data types showing the Github repository \href{https://github.com/rrousselGit/flutter\_hooks}{rrousselGit/flutter\_hooks} with file \href{https://github.com/rrousselGit/flutter\_hooks/blob/master/packages/flutter\_hooks/test/use_automatic\_keep\_alive\_test.dart}{use\_automatic\_keep\_alive\_test.dart}.
        Note that we omit line, word and character annotations in (b) for better readability.
    }
    \label{fig:data:types}
    \vspace{-1cm}
\end{figure}

We utilize the HTML and source code file to automatically generate all annotations.
The bounding box of any IDE element can be retrieved by manually identifying the CSS selector for the relevant element inside the HTML and accessing its bounding box information.
Since the naming is consistent across all HTML files, this process is necessary only once per element.
We annotate numerous IDE elements (e.g. coding area, active tab, status bar etc.) and refer to \cref{tab:det} for a full list and to \cref{fig:data:types:bbox} for a visualization.
Note, that arbitrary additional annotations for other IDE elements can be added any time to our dataset.
By taking an automated screenshot of the scene using the original resolution, we get the underlying image to which the annotations can be applied directly without any offsets.

Moreover, we extract annotations for text detection and recognition for the source code of the active file visible in the coding area.
This is done on a per-line and per-word basis.
Additionally, we provide annotations for the \textit{coding grid}, i.e. the tight bounding box around the visible code including the character height and width.
These annotations of only six numbers (four for the bounding box and character width and height) enables the computation of the coding grid as visualized in \cref{fig:data:types:grid}.
The coding grid can be used to retrieve line and character-based text bounding boxes and brings the additional benefit, that it is much easier to accurately retrieve indentation, which is crucial for languages such as Python.

Since we additionally analyze the influence of binarization, i.e. conversion to a clean black-and-white version of the screenshot, we automatically generate annotations for this use case.
This is done by creating a binarized HTML version of the scene by identifying the relevant CSS selectors for source code text.
We enforce black color for all source code text and white for everything else by adjusting the CSS parameters.
See \cref{fig:data:types:bw} for an example.

The dataset is split into a training, validation and test set of sizes $7,561$, $1,921$, and $2,518$, respectively.
There is no overlap of fonts across the splits, and only a minor overlap across themes, which was inevitable due to the previously mentioned issues in automating the theme selection process.

\section{Evaluation}
\label{sec:eval}

Coding screen cast analysis comprises several parts.
For our evaluations, we assume videos are processed frame-by-frame and thus, consider only single frames as input for the respective components.
We evaluate IDE element detection in \cref{sec:eval:det}.
Next, we evaluate image binarization in \cref{sec:eval:bin}, which can help to improve text recognition performance as analyzed in \cref{sec:eval:ocr}.
In addition to the influence of binarization, we also compare different OCR engines and investigate how they are affected by changes in the input image quality.

\subsection{Object Detection}
\label{sec:eval:det}

Since the focus of this work is not to improve object detection algorithms, we benchmark our dataset on a well-established baseline.
We use a Mask R-CNN \cite{heMaskRCNN2017} with a ResNet-50-FPN \cite{heDeepResidualLearning2016,linFeaturePyramidNetworks2017} backbone that was pre-trained on MS COCO \cite{linMicrosoftCOCOCommon2014} for our experiments.
We freeze the weights at stage four and use stochastic gradient descent with momentum, a batch size of 16 and a cosine learning rate schedule \cite{loshchilovSGDRStochasticGradient2017}.
For the learning rate schedule, we set the initial learning rate to $0.001$, the final learning rate to $0$ after $10,000$ and use a linear warm-up during the first $1,000$ iterations.
The results are summarized in \cref{tab:det}.
Some elements, such as \textit{sidebar}, \textit{tabs and actions container}, \textit{minimap} and \textit{output panel} achieve a Box AP above $80$.
Other elements, such as \textit{title bar}, \textit{notifications}, \textit{last code line number}, \textit{suggestion widget}, \textit{tabs breadcrumbs} are harder to detect and the Box AP lies below $50$.
The bounding box of the \textit{coding grid} and the ones for source code text \textit{lines} are most important for text detection and recognition, and achieve a Box AP of $69.9$ and $71.9$, respectively.

\begin{table}[h]
    \centering
    \begin{tabular}{lccc}
        \toprule
                                         & \multicolumn{3}{c}{Bounding Box}                       \\
        Class name                       & ~~AP~~                           & ~~AP50~~ & ~~AP75~~ \\
        \midrule
        sidebar                          & 92.0                             & 99.8     & 98.7     \\
        minimap                          & 84.6                             & 99.0     & 97.6     \\
        tabs\_and\_actions\_container~~~ & 82.9                             & 98.2     & 96.4     \\
        output\_panel                    & 82.4                             & 97.6     & 93.9     \\
        status\_bar                      & 79.5                             & 99.0     & 95.4     \\
        activity\_bar                    & 76.8                             & 95.6     & 82.0     \\
        editor\_container                & 72.5                             & 92.4     & 84.7     \\
        line                             & 71.9                             & 93.3     & 87.4     \\
        code\_container                  & 71.9                             & 90.9     & 82.9     \\
        coding\_grid                     & 69.9                             & 96.7     & 78.6     \\
        code\_line\_number\_bar          & 62.1                             & 88.3     & 66.9     \\
        scrollbar                        & 56.3                             & 80.0     & 70.1     \\
        active\_tab                      & 53.1                             & 70.0     & 60.0     \\
        find\_widget                     & 53.1                             & 69.6     & 63.8     \\
        last\_code\_line\_number         & 48.7                             & 80.7     & 56.2     \\
        notifications                    & 48.4                             & 89.2     & 40.6     \\
        suggestion\_widget               & 48.2                             & 68.9     & 57.4     \\
        tabs\_breadcrumbs                & 45.4                             & 71.7     & 53.0     \\
        title\_bar                       & 31.6                             & 44.5     & 40.2     \\
        \bottomrule
    \end{tabular}
    \vspace{5mm}
    \caption{
        Quantitative object detection performance per class using Box AP.
    }
    \label{tab:det}
    \vspace{-1cm}
\end{table}

\subsection{Binarization}
\label{sec:eval:bin}

\begin{figure}[h]
    \centering
    \begin{subfigure}[b]{\linewidth}
        \fbox{\includegraphics[width=0.48\linewidth]{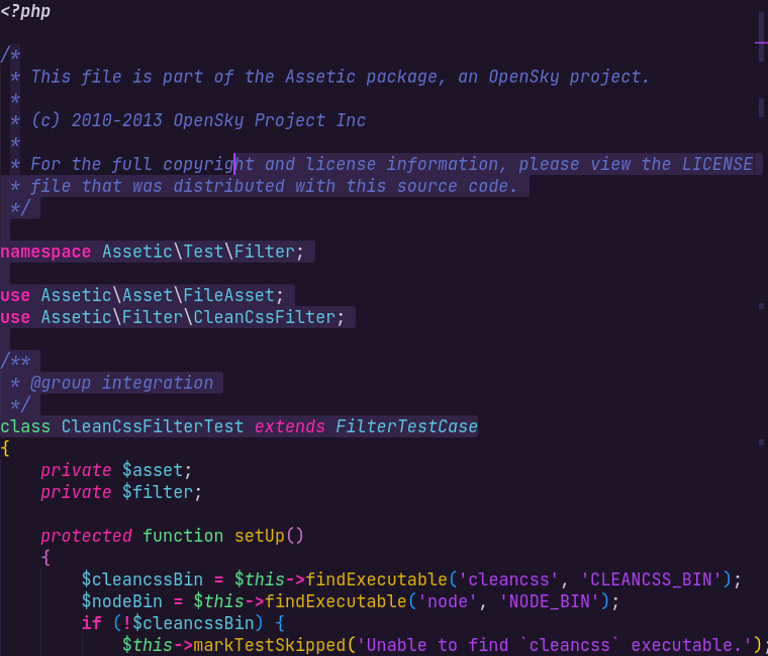}}
        \fbox{\includegraphics[width=0.48\linewidth]{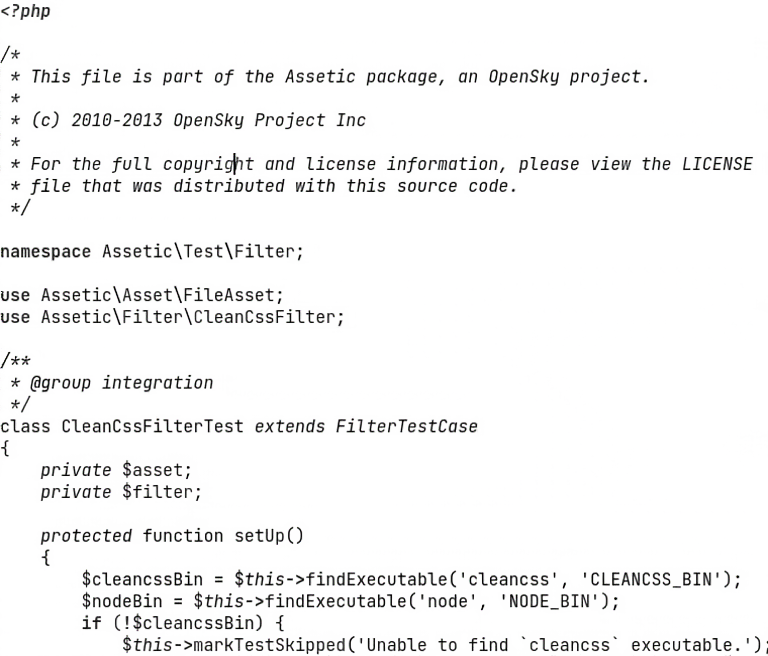}}
    \end{subfigure}
    \caption{
        Qualitative examples of the binarization showing the input image on the left and the predicted binarization on the right.
    }
    \label{fig:eval:bin:ex}
    \vspace{-1cm}
\end{figure}

\begin{figure}[h]
    \begin{subfigure}[b]{\linewidth}
        \fbox{\includegraphics[width=0.48\linewidth]{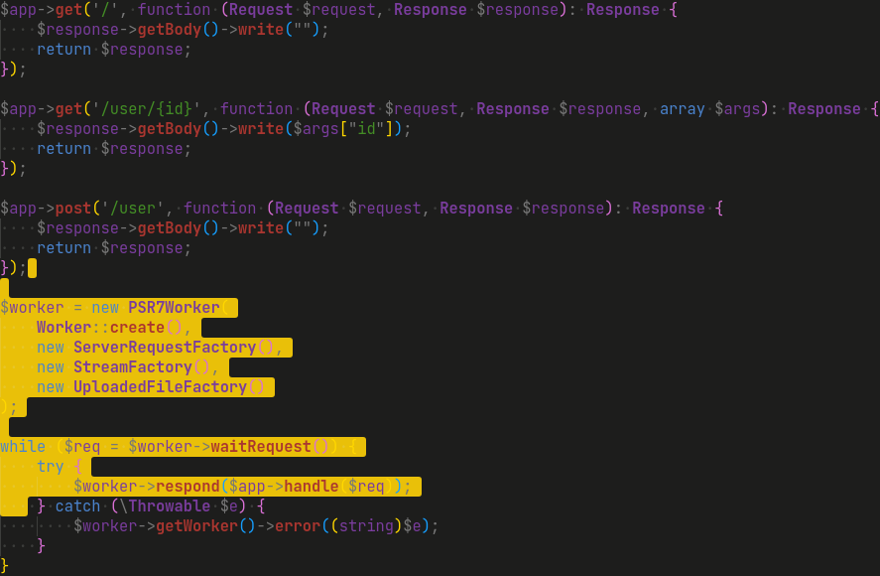}}
        \fbox{\includegraphics[width=0.48\linewidth]{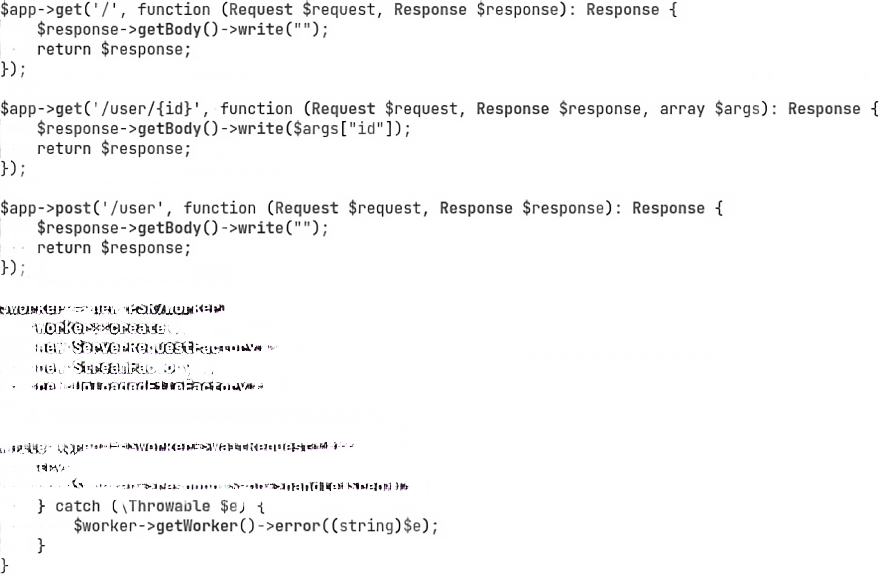}}\\
        ~\\%
        \fbox{\includegraphics[width=0.48\linewidth, trim={0 200px 0 0px}, clip]{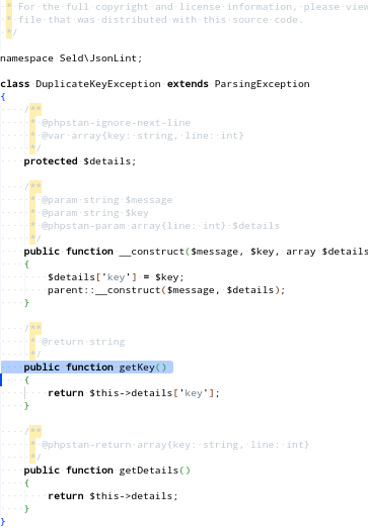}}
        \fbox{\includegraphics[width=0.48\linewidth, trim={0 200px 0 0px}, clip]{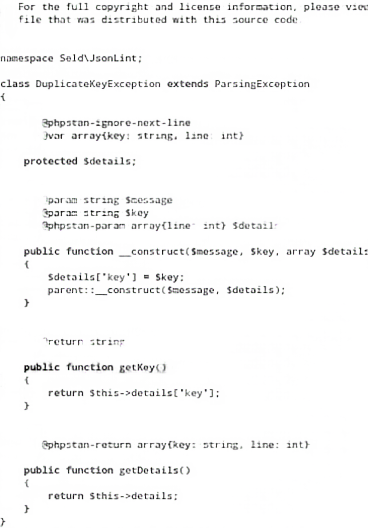}}
    \end{subfigure}
    \caption{
        Challenging qualitative examples of the binarization showing the input image on the left and the predicted binarization on the right.
    }
    \label{fig:eval:bin:diff}
\end{figure}

We investigate the performance of image binarization approaches for converting a screenshot from a coding screencast to a binarized black-and-white version.
For our experiments, we train Pix2PixHD \cite{wangHighResolutionImageSynthesis2018}, the successor of the very popular Pix2Pix \cite{isolaImagetoImageTranslationConditional2017} architecture for image-to-image translation.
We train for 40 epochs and leave the remaining original training configuration unchanged.
The results are summarized in \cref{tab:bin} and we provide qualitative samples in \cref{fig:eval:bin:ex,fig:eval:bin:diff}.
Overall, binarization works well as indicated by the mean Peak Signal-to-Noise Ratio (PSNR) of $5.49$ in \cref{tab:bin} and the qualitative example in \cref{fig:eval:bin:ex}.
The model seems to have difficulties especially with rare strong contrasts and very low contrast as indicated by the examples in \cref{fig:eval:bin:diff}.

\begin{table}[h]
    \centering
    \begin{tabular}{lcccc}
                        & ~~Accuracy $\uparrow$~~ & ~~PSNR $\uparrow$~~ & ~~DRDM $\downarrow$~~ \\
        \toprule
        Median          & 72.07                   & 5.54                & 67.52                 \\
        25\% Quantile~~ & 67.09                   & 4.82                & 62.74                 \\
        75\% Quantile   & 76.01                   & 6.20                & 73.46                 \\
        Mean            & 70.81                   & 5.49                & 165.70                \\
        Std.            & 7.77                    & 1.08                & 1699.39               \\
        \bottomrule
    \end{tabular}
    \vspace{5mm}
    \caption{
        Quantitative binarization performance.
    }
    \label{tab:bin}
    \vspace{-1cm}
\end{table}

\subsection{Optical Character Recognition}
\label{sec:eval:ocr}

We benchmark five different approaches for text recognition in the following.
For the evaluation, we create a subset of our test split by randomly sampling 10 words from each of the 2,518 test images to reduce the computational workload while maintaining statistical significance.
To eliminate the influence of text detection performance, we use the ground truth bounding boxes.
We investigate the influence of binarization (color image vs. ground truth binarization vs. predicted binarization) and image quality (20\%-100\% of original image size as input) on text recognition performance.
The state-of-the-art text recognizers we benchmark are:
\begin{itemize}
    \item Tesseract \cite{smithOverviewTesseractOCR2007a} (common baseline)
    \item SAR \cite{liShowAttendRead2019} pre-trained mainly on MJSynth \cite{jaderbergDeepFeaturesText2014,jaderbergReadingTextWild2016}, SynthText \cite{guptaSyntheticDataText2016} and SynthAdd \cite{liShowAttendRead2019}
    \item MASTER \cite{luMASTERMultiaspectNonlocal2021} pre-trained on MJSynth \cite{jaderbergDeepFeaturesText2014,jaderbergReadingTextWild2016}, SynthText \cite{guptaSyntheticDataText2016} and SynthAdd \cite{liShowAttendRead2019}
    \item ABINet \cite{fangReadHumansAutonomous2021} pre-trained on  MJSynth \cite{jaderbergDeepFeaturesText2014,jaderbergReadingTextWild2016} and SynthText \cite{guptaSyntheticDataText2016} %
    \item TrOCR \cite{liTrOCRTransformerBasedOptical2023} fine-tuned on the SROIE dataset \cite{huangICDAR2019CompetitionScanned2019} \footnote{See \href{https://rrc.cvc.uab.es/?ch=13&com=introduction}{https://rrc.cvc.uab.es/?ch=13}}
\end{itemize}

Note, that TrOCR does not distinguish lowercase and capital letters.
Thus, we convert the ground truth to lowercase for computing the mean Character Error Rate (CER) in only this case.
This obviously simplifies the task, which is why these results cannot be fairly compared with the other approaches.

On the full image resolution, we can observe from \cref{tab:eval:ocr} that TrOCR and Tesseract benefit most from having access to the ground truth or predicted binarized version of the input image.
Overall best performance is $0.08$ mean CER for TrOCR and the ground truth binarized input images.
The performance of SAR, MASTER and ABINet is more robust on color images which means those approaches do not benefit from a prior conversion to a black-and-white image using Pix2PixHD.
We argue that these differences stem from the differences in training data.
In scenarios where mostly binarized training data is available, using image-to-image translation approaches is a valid and helpful way to boost performance on the final task.

\begin{table*}
    \centering
    \begin{tabular}{lccc}
        \toprule
        ~~CER~$\downarrow$~~                                 & ~~Binarized (GT)~~   & ~~Binarized (Pred.)~~ & ~~Color~~            \\
        \midrule
        ABINet \cite{fangReadHumansAutonomous2021}           & 0.37 (0.59)          & 0.43 (0.78)           & 0.43 (0.94)          \\
        Tesseract \cite{smithOverviewTesseractOCR2007a}      & 0.32 (0.45)          & 0.41 (0.46)           & 0.67 (0.51)          \\
        SAR \cite{liShowAttendRead2019}                      & 0.26 (0.73)          & 0.35 (0.97)           & 0.28 (0.86)          \\
        MASTER \cite{luMASTERMultiaspectNonlocal2021}~~      & \textbf{0.24 (0.95)} & \textbf{0.35 (0.97)}  & \textbf{0.25 (0.99)} \\
        \midrule
        TrOCR$^*$  \cite{liTrOCRTransformerBasedOptical2023} & \textit{0.08 (0.27)} & \textit{0.14 (0.33)}  & \textbf{0.25 (0.93)} \\
        \bottomrule
    \end{tabular}
    \vspace{5mm}
    \caption{
        Quantitative results at full image resolution using the binarized, converted (from \cref{sec:eval:bin}) and color image.
        We report the \textit{mean (standard deviation)} for the Character Error Rate (CER).
        ($^*$ CER is computed ignoring capitalization)
    }
    \label{tab:eval:ocr}
\end{table*}

All approaches show robust performance when the image quality is degraded slightly, i.e. when it is reduced to 60\%-70\% of the original image size.
This effect is constant across image types as seen in Fig. 5. %

\begin{figure}[h]
    \centering
    \begin{subfigure}[b]{0.5\linewidth}
        \centering
        \includegraphics[width=\linewidth]{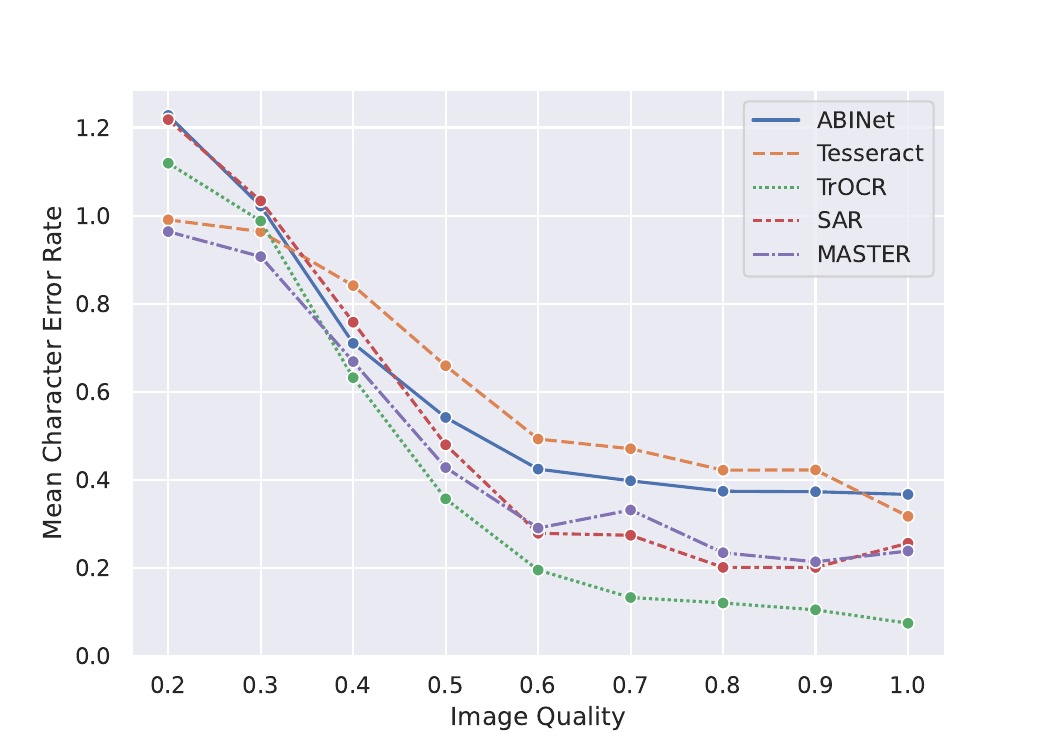}
        \caption{
            Binarized (GT) Images
        }
        \label{fig:eval:ocr:bin}
    \end{subfigure}~%
    \begin{subfigure}[b]{0.5\linewidth}
        \centering
        \includegraphics[width=\linewidth]{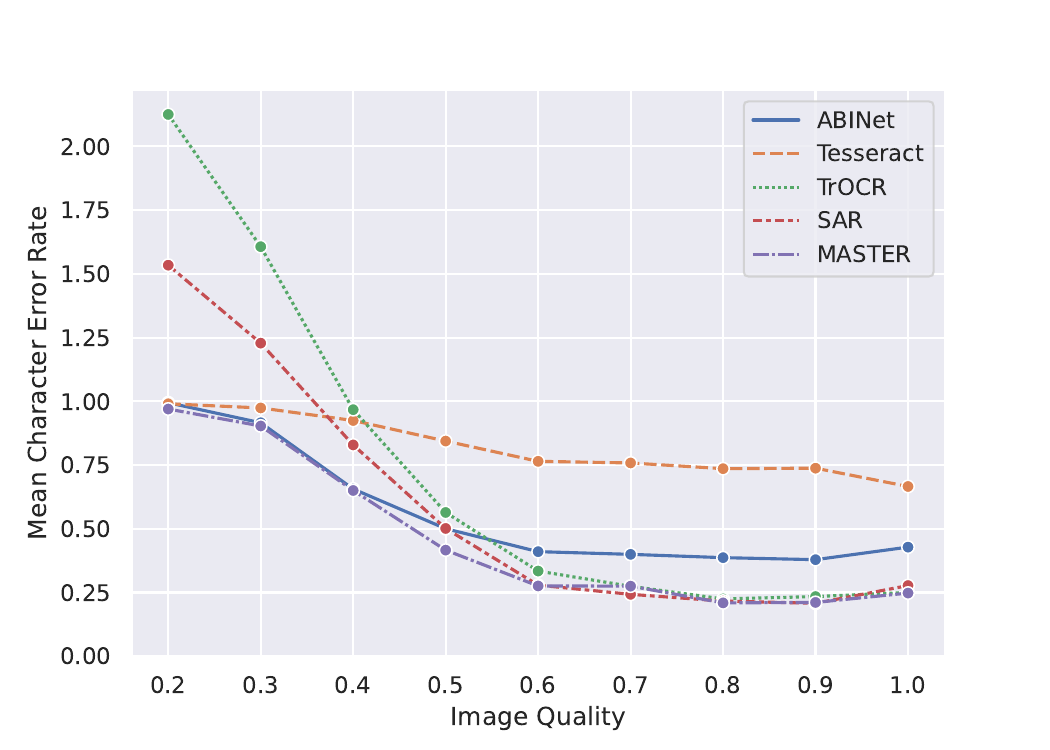}
        \caption{
            Binarized (Pred.) Images
        }
        \label{fig:eval:ocr:color}
    \end{subfigure}
    \begin{subfigure}[b]{0.5\linewidth}
        \centering
        \includegraphics[width=\linewidth]{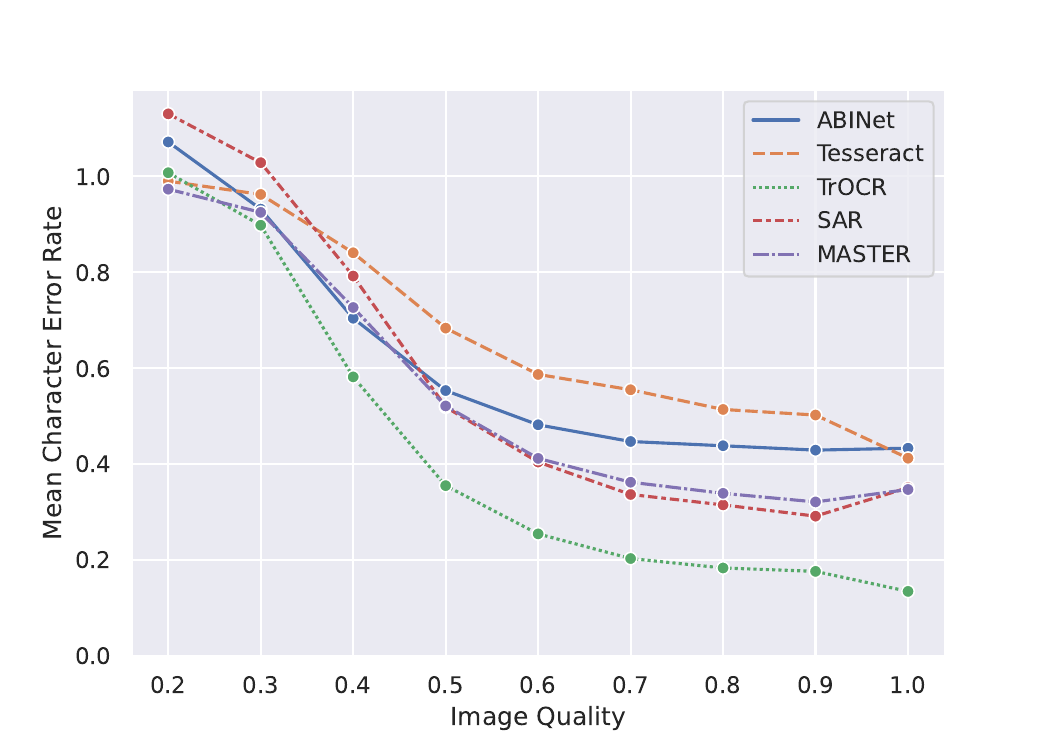}
        \caption{
            Color Images
        }
        \label{fig:eval:ocr:pred}
    \end{subfigure}
    \caption{
        Quantitative evaluation of OCR performance for different image qualities.
    }
    \label{fig:eval:ocr}
\end{figure}

\section{Conclusion}
\label{sec:conclusion}

Coding screencasts have emerged as powerful tools for programming education of novices as well as experienced developers.
In addition to the video content, having access to the full source code of the project at any given time of the programming tutorial brings two major benefits:
(1) platform-wide and tutorial-based search can leverage the full source code to enhance the precision and relevance of search results and (2) it empowers learners to engage in hands-on experimentation with the source code seamlessly.
In this work, we work towards enabling detailed information retrieval from such coding screencasts.
We present a novel high-quality dataset called \codescan{}, comprising $12,000$ fully annotated IDE screenshots.
\codescan{} is highly diverse and contains 24 programming languages, over 90 different themes of \vscode{}, and 25 fonts while at the same time varying the IDE appearance through changing the visibility, position and size of different IDE elements (e.g. sidebar, output panel).
Our evaluations show that baseline object detectors are suitable for text recognition and achieve a Box AP for source code line detection of $71.9$.
Moreover, we showed that baseline image-to-image translation architectures are well suited for coding screencast image binarization.
Since we used baseline architectures for object detection and image binarization, we expect that performance can be increased significantly by resorting to state-of-the-art models.
Finally, we compared different OCR engines and analyzed their dependence on image quality and image binarization.
We found that binarization can boost performance for some OCR engines, while slight quality degradations in terms of image resolution do not significantly affect the text recognition quality.
In future work, we plan to tackle the full source code retrieval from programming videos pipeline by leveraging our introduced \codescan{} dataset.

Several future directions are very interesting.
Since we limit ourselves to screenshot analysis, the tracking, composing and synchronizing files during a video tutorial remains an open task.
In addition, evaluation metrics could move from classical text recognition towards evaluating code executability and the tree distance to the original abstract syntax tree.
Finally, the coding grid parameters could be estimated using an additional \textit{coding grid head}.
Its availability is expected to significantly simplify the identification of the correct indentation.

\printbibliography

\end{document}